\documentclass[letterpaper, 10 pt, conference]{ieeeconf}  % Comment this line out if you need a4paper

\IEEEoverridecommandlockouts                              % This command is only needed if 
\usepackage[utf8]{inputenc} % allow utf-8 input
\usepackage[T1]{fontenc}    % use 8-bit T1 fonts
\usepackage{hyperref}       % hyperlinks
\usepackage{url}            % simple URL typesetting
\usepackage{booktabs}       % professional-quality tables
\usepackage{amsfonts}       % blackboard math symbols
\usepackage{nicefrac}       % compact symbols for 1/2, etc.
\usepackage{microtype}      % microtypography
\usepackage{xcolor}         % colors

\usepackage{amsmath}
\usepackage{graphicx}
\usepackage{multirow}
\usepackage{subcaption}
\usepackage{tabularx}
\usepackage{wrapfig}
\usepackage{amssymb}
\usepackage{fancyvrb} % control verbatim size

\usepackage{xspace}

\usepackage{multirow}
\usepackage{colortbl}
\usepackage{booktabs}
\usepackage{multicol}
\usepackage{tcolorbox}
\usepackage{xspace}
\usepackage{epigraph}

\usepackage{lmodern}
\usepackage[T1]{fontenc}
\usepackage{svg}

\newcommand{\modelname}{MorphoSim\xspace}

\title{\LARGE \bf
MorphoSim: An Interactive, Controllable, and Editable Language-guided 4D World Simulator
}

\author{Xuehai He$^{*1}$,
    Shijie Zhou$^{*2}$,
    Thivyanth Venkateswaran$^{3}$,
    Kaizhi Zheng$^{1}$, \\
    Ziyu Wan$^{4}$,
    Achuta Kadambi$^{2}$,
    Xin Eric Wang$^{1}$\\[2mm]
    $^{1}$University of California, Santa Cruz, $^{2}$University of California, Los Angeles,  $^{3}$ IIT Bombay, $^{4}$ Microsoft\\[1mm]
    \texttt{\{xhe89, xwang366\}@ucsc.edu}\thanks{* equal contribution}
}

\begin{document}

\maketitle
\thispagestyle{empty}
\pagestyle{empty}

%%%%%%%%%%%%%%%%%%%%%%%%%%%%%%%%%%%%%%%%%%%%%%%%%%%%%%%%%%%%%%%%%%%%%%%%%%%%%%%%
\begin{figure*}[t]
    \centering
    \includegraphics[width=\linewidth]{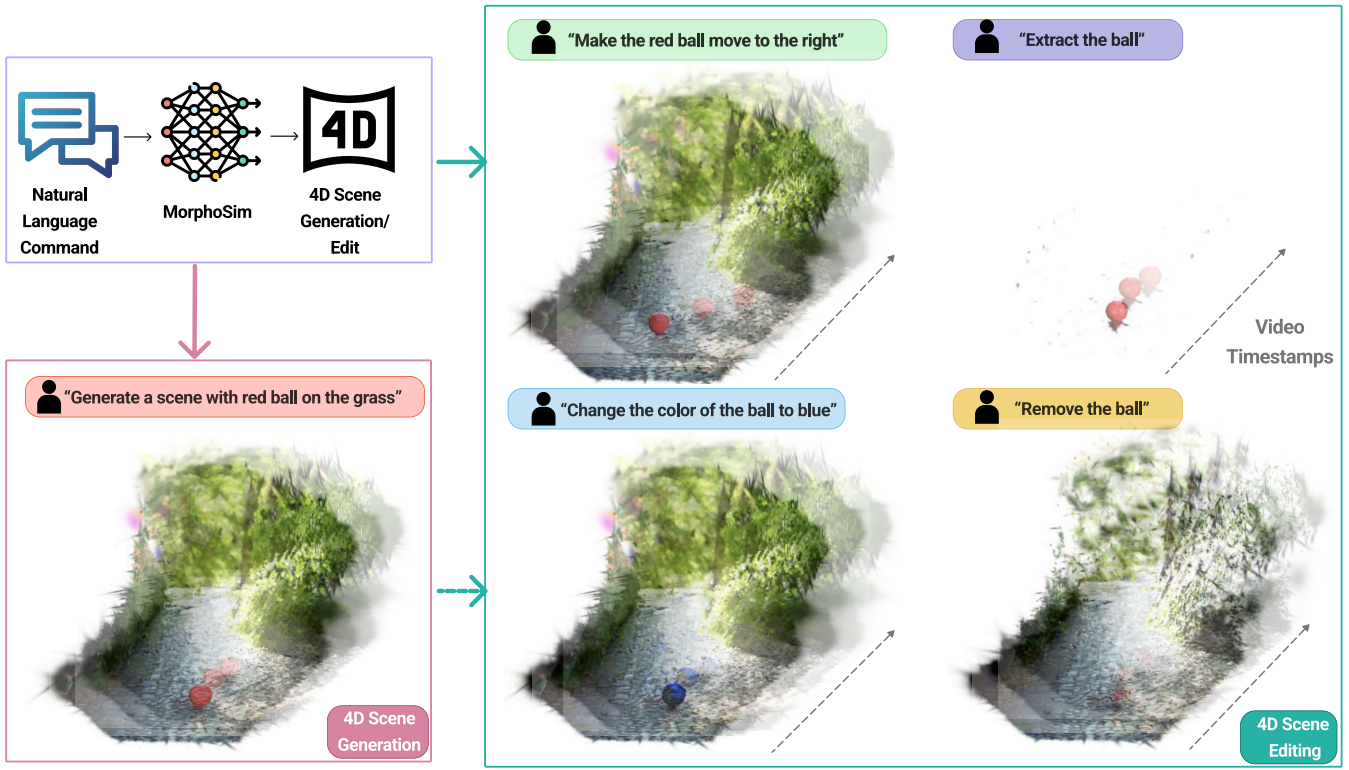}
    \caption{\modelname is a fully natural language-guided 4D scene generation engine that enables generation and editing of 4D scenes based on language commands. Given a natural language input, \modelname constructs a 4D scene and provides a unified framework for multiple tasks, including high-quality scene generation, interactive modification of object motion and appearance, and object extraction or removal.}
    \label{fig:teaser}
\end{figure*}

% \begin{figure}[t]
%     \centering
%     \includesvg[width=\linewidth]{figures/teaser1.svg} % no .svg extension needed
%     \caption{\modelname is a fully natural language-guided 4D scene generation engine that enables generation and editing of 4D scenes based on language commands. Given a natural language input, \modelname constructs a 4D scene and provides a unified framework for multiple tasks, including high-quality scene generation, interactive modification of object motion and appearance, and object extraction or removal.}
%     \label{fig:teaser}
% \end{figure}

\begin{abstract}
World models that support controllable and editable spatiotemporal environments are valuable for robotics, enabling scalable training data, reproducible evaluation, and flexible task design. While recent text-to-video models generate realistic dynamics, they are constrained to 2D views and offer limited interaction. We introduce MorphoSim, a language-guided framework that generates 4D scenes with multi-view consistency and object-level controls. From natural language instructions, MorphoSim produces dynamic environments where objects can be directed, recolored, or removed, and scenes can be observed from arbitrary viewpoints. The framework integrates trajectory-guided generation with feature field distillation, allowing edits to be applied interactively without full re-generation. Experiments show that MorphoSim maintains high scene fidelity while enabling controllability and editability. The code is available at~\url{https://github.com/eric-ai-lab/Morph4D}.
\end{abstract}

\section{Introduction}
\label{sec:intro}
Robotics needs world models that support observation from many viewpoints, evolve over time, and accept direct intervention for task specification, data generation, and evaluation. Recent text-to-video models~\cite{kling2023,nvidia2025cosmos,chen2024videocrafter2,wan2.1,kong2024hunyuanvideo} show that large generative models can produce high quality dynamics from language prompts, and there is growing interest in using such models as simulators~\cite{vidgen_worldmodel,wang2023world} for robotic environment. Yet most systems remain 2D and single-view, primarily relying on diffusion models or autoregressive models~\cite{xiang2024pandora,wang2023world,tian2024visual}, restricting them to single-view observations and non-interactive simulations. These limitations prevent such models from accurately capturing the true complexity of dynamic, multi-view environments. They also do not expose controls that robot learning needs: camera view control, object-level motion control, object insertion or removal, and interactive edit operations.

This paper targets a language-guided~\emph{4D} (space–time) simulator that supports multi-view rendering and interactive editing at the object level. We ask two questions that are central for robotics: (i) can we lift the capability of visual generation models to 4D (spatial-temporal) scenarios? and (ii) can we expose language-driven controls that change object trajectories and appearance so that one can script tasks, perturb scenes, and generate diverse training data for visuomotor policies?

We present \modelname, a language-guided world simulator that converts natural language commands into editable 4D scenes with consistent multi-view dynamics. By supporting interactive control and object-level editing, \modelname enables robotics applications that require flexible scene variation. In particular, it can generate synthetic training data for policy learning, provide controlled perturbations for closed-loop evaluation, and support rapid construction of task variants for studying long-horizon planning. The ability to edit object motion and appearance also facilitates robustness testing of perception systems under viewpoint changes, occlusions, and counterfactual scene modifications. For example, given the instruction “a red cube moves to the plate while the camera circles the table; then make the cube blue and reverse its motion,” \modelname produces a temporally coherent, multi-view sequence and applies the specified edits without re-generating the entire scene.

Three challenges arise. \underline{First}, a suitable embodied scene representation under this situation is not well defined, it must support consistent geometry, appearance, and motion under arbitrary viewpoints. \underline{Second}, standard text-to-video backbones are optimized for single-view synthesis and do not maintain multi-view coherence or camera control. \underline{Third}, robotics needs object-level handles (velocity, color, presence) that can be bound to language instructions and edited interactively.

\modelname addresses these points with a modular design. A command parameterizer parses language into structured controls for camera and objects; A scene generator produces a 4D representation that supports view-consistent rendering; A scene editor exposes two editing paths: a dynamic control submodule that steers object motion directions and trajectories from language, and a static edit submodule that changes object appearance, extracts objects, or removes them from the scene. The editor operates directly on the 4D representation so that edits are fast and preserve temporal and multi-view consistency.

Our contributions are as follows:
\begin{itemize}
\item We present \modelname, a new language-guided simulator framework with three components: a \textit{command parameterizer}, a \textit{scene generator}, and a \textit{scene editor}. Given natural language, it produces view-consistent 4D scenes and supports interactive edits through language.
\item The scene editor includes two novel submodule designs: a \textit{Dynamic Control} submodule that changes object motion directions and trajectories from language commands, and a \textit{Static Edit} submodule that changes object appearance (e.g., color), extracts objects, or removes objects.
\item We evaluate \modelname on robotics-oriented scenarios. Results show high-fidelity 4D scenes and effective control and edit operations that support synthetic data generation and controlled evaluation, with improvements over representative baselines in controllability and editability.
\end{itemize}

\begin{figure*}[t]
  \centering
%   \fbox{\rule{0pt}{3in} \rule{0.9\linewidth}{0pt}}
   \includegraphics[width=\linewidth]{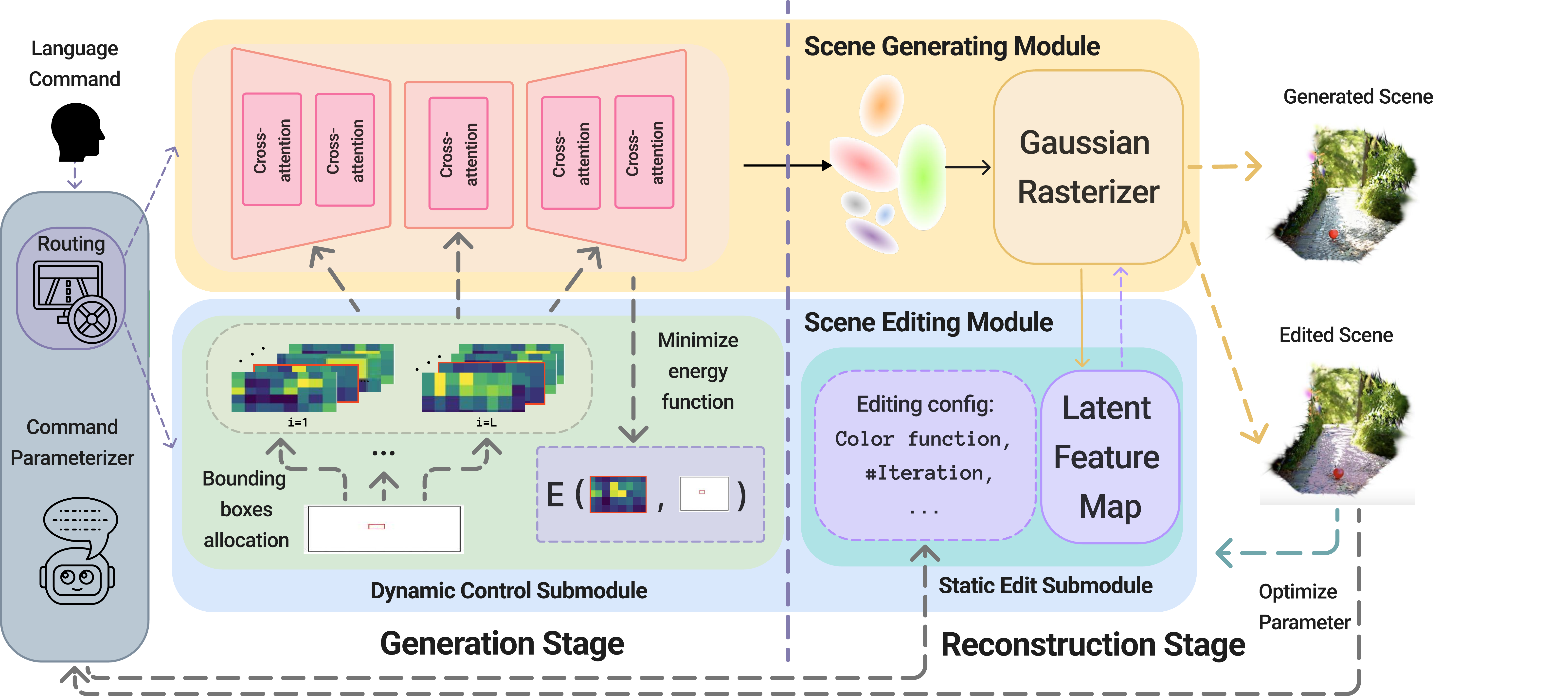}
   \caption{\textbf{Overview of the~\modelname pipeline}. It consists of a \emph{command parameterizer} for natural language comprehension, a controllable \emph{scene generating module} which supports generation of 4D scenes following dynamic objects motion guidance, and an interactive \emph{scene editing module} for executing edits. }
   \label{fig:overall}
\end{figure*}

% \iffalse
\section{Related Works}
\noindent \textbf{Generative Models for 4D Scenes} Recent works in diffusion models have revolutionized visual generation across 2D, 3D, and 4D domains. In 2D, frameworks such as Stable Diffusion~\cite{stable_diffusion} and Imagen~\cite{imagen2022} enable high-fidelity text-to-image synthesis and retrieval~\cite{he2023discffusion}, while cascaded and latent diffusion approaches extend these capabilities to text-to-visual generation~\cite{latentvideodiffusion,easyanimate2024xu,opensora,he2024flexecontrol,opensoraplan,training-free-guidance,yang2024cogvideox,vchitect2024}. Building upon these successes, diffusion models have transformed visual generation in 3D space~\cite{liu2024sherpa3d,feng2023layoutgpt,chen20243dtopia,wu2024direct3d}. Parallel to these advances,  Diffusion priors have further driven 4D scene generation~\cite{singer2023text,yu20254real,bahmani2024tc4d,zhao2023animate124,bahmani20244d,ling2024align,ren2023dreamgaussian4d,xu2024comp4d,yin20234dgen,ren2024l4gm,li20244k4dgen,sun2024eg4d}, producing realistic spatiotemporal content under constrained settings. Parallel to these advances, efficient scene representations such as 3D Gaussian Splatting (3DGS)~\cite{kerbl20233d} have enabled high-fidelity 3D reconstruction and view synthesis, with dynamic extensions~\cite{wu20234d,duan20244d,yang2023real,huang2024sc,liang2023gaufre,wang2024shape,lei2024mosca,mihajlovic2025splatfields,stearns2024dynamic} modeling temporal deformation fields for 4D consistency. Diffusion priors have further driven 4D scene generation~\cite{singer2023text,yu20254real,bahmani2024tc4d,zhao2023animate124,bahmani20244d,ling2024align,ren2023dreamgaussian4d,xu2024comp4d,yin20234dgen,ren2024l4gm,li20244k4dgen,sun2024eg4d}, producing realistic spatiotemporal content under constrained settings. For robotics, however, 4D world models must not only render scenes but also support control, editing, and task-driven variation. Our work addresses this gap by introducing a unified, language-guided framework that combines high-quality 4D generation with interactive editing capabilities.

\noindent \textbf{Language-guided Scene Editing} 
Recent advances in language-guided scene editing have shown promising results in static 3D settings by leveraging neural implicit representations and 2D diffusion priors for tasks such as appearance modification~\cite{chen2024gaussianeditor,wang2024gaussianeditor,zhuang2023dreameditor,khalid2024latenteditor,sella2023vox,xu2024gg,fang2024chat,wu2024gaussctrl}, object replacement~\cite{bartrum2024replaceanything3d,zheng2025editroom,zhuang2024tip,gordon2023blended}, and removal~\cite{zhuang2023dreameditor,gaussian_grouping,qiu2024feature}. In the realm of 4D editing, methods like 4D-Editor~\cite{jiang20234d}, CTRL-D~\cite{he2024ctrl}, Control4D~\cite{shao2024control4d}, and Instruct 4D-to-4D~\cite{mou2024instruct} extend these techniques to dynamic scenes by ensuring temporally consistent appearance edits across frames. Our work advances this line of research by introducing motion control through natural language instructions, thereby enabling not only appearance modifications but also explicit edits to the underlying motion patterns in scenes.

\section{\modelname}
\modelname is a unified framework that integrates multiple functionalities for generating and editing 4D scenes from natural language input. As illustrated in Figure~\ref{fig:overall}, \modelname consists of three core modules: the \textit{Command Parameterizer} Module, the \textit{Scene Generation Module}, and the \textit{Scene Editing Module}. The Command Parameterizer Module serves as the interface between natural language input and system execution. It interprets user instructions, routes them to the appropriate module, and converts them into structured, executable commands. The Scene Generation Module is responsible for generating dynamic scenes based on language descriptions, capturing spatial and temporal instructions. The Scene Editing Module enables interactive modifications, allowing users to adjust motion trajectories, alter object appearances (color), and manipulate scene elements (delete or extract) through natural language instructions. In the following sections, we will introduce these modules in detail.

\subsection{Preliminaries}
\noindent \textbf{Video Diffusion Models}
\label{sec:preliminaries}
Video Diffusion Models (VDMs)~\cite{latentvideodiffusion,stable_video_diffusion} extend diffusion models to video generation by formulating a fixed forward diffusion process that progressively corrupts a 4D video sample $\mathbf{x}_0$ with noise in the latent space. This enables the model to learn a reverse denoising process to recover the original video. 

The forward diffusion process consists of $T$ timesteps, where noise is gradually added to the clean data $\mathbf{x}_0$ through a Markovian parameterization:
\begin{equation}
q(\mathbf{x}_t \mid \mathbf{x}_{t-1}) = \mathcal{N}(\mathbf{x}_t; \sqrt{1 - \beta_t} \mathbf{x}_{t-1}, \beta_t \mathbf{I}),
\end{equation}
\begin{equation}
q(\mathbf{x}_t \mid \mathbf{x}_0) = \mathcal{N}(\mathbf{x}_t; \sqrt{\bar{\alpha}_t} \mathbf{x}_0, (1 - \bar{\alpha}_t) \mathbf{I}),
\end{equation}
where $\beta_t$ is a predefined variance schedule during diffusion sampling, $\alpha_t = 1 - \beta_t$, and $\bar{\alpha}_t = \prod_{i=1}^{t} \alpha_i$. 
The reverse process then attempts to reconstruct $\mathbf{x}_{t-1}$ from $\mathbf{x}_t$ by learning a denoising distribution:
\begin{equation}
p_{\theta} (\mathbf{x}_{t-1} \mid \mathbf{x}_t) = \mathcal{N} (\mathbf{x}_{t-1}; \boldsymbol{\mu}_{\theta} (\mathbf{x}_t, t), \boldsymbol{\Sigma}_{\theta} (\mathbf{x}_t, t)).
\end{equation}
Here, the mean $\boldsymbol{\mu}_{\theta}$ and variance $\boldsymbol{\Sigma}_{\theta}$ are the estimated Gaussian
mean and variance predicted by the denoising network. The update step is typically computed as:
\begin{equation}
\mathbf{x}_{t-1} = \frac{1}{\sqrt{\alpha_t}} \left( \mathbf{x}_t - \sqrt{1 - \alpha_t} \epsilon_\theta(\mathbf{x}_t, t, \mathbf{c}) \right) + \sigma_t \mathbf{z},
\end{equation}
where $\alpha_t$ is the noise schedule coefficient, $\sigma_t$ is the stochastic noise factor, and $\mathbf{z} \sim \mathcal{N}(0, I)$ represents Gaussian noise injected at each step for improved sample diversity. The latent feature update can be modified to control the generation direction.

\noindent \textbf{3D Gaussian Splatting}
In dynamic scene reconstruction approaches~\cite{wang2024shape, lei2024mosca, stearns2024dynamic}, the scenes was represented with dynamic 3D Gaussians~\cite{luiten2023dynamic}—a set of persistent 3D Gaussians~\cite{kerbl20233d} that deform over time to model motion. These representations cann efficiently capture spatiotemporal variations in monocular video reconstructions. To created 4D scenes, we build upon MoSca~\cite{lei2024mosca}, which resconstruct 4D scenes with single-view partial observations by leveraging priors from 2D foundation models~\cite{unidepth,zoedepth,teed2020raft,cotracker,harley2022particle} and by imposing regularization constraints on Gaussian motion trajectories. The key in the techniques is the use of 4D Motion Scaffold, a structured graph $(\mathcal{V}, \mathcal{E})$ that governs the deformation of individual 3D Gaussians $\mathcal{G} = \{G_j\}_{j=1}^n$.

\subsection{Language-Guided 4D Scene Generation}
\subsubsection{LLM as a Command Parameterizer}

Given an input natural language instruction $\mathcal{L}$, \modelname employs a large language model (LLM) agent $\mathcal{A}$ to interpret the command, extract semantic attributes, and dynamically route the request to the appropriate execution module. The agent $\mathcal{A}$ formalizes this routing process by mapping the input $\mathcal{L}$ to an execution plan $\mathcal{P}$: $\mathcal{A}: \mathcal{L} \to \mathcal{P} = (\mathcal{M}, \mathcal{Q}),$ where $\mathcal{M} \in \{\text{GEN}, \text{EDIT}\}$ represents the routing decision, selecting either the scene generating module ($\mathcal{G}$) or the scene editing module ($\mathcal{E}$), and $\mathcal{Q}$ represents a set of structured queries extracted from the input.

\subsubsection{Scene Generating Module}

If the LLM agent $\mathcal{A}$ determines $\mathcal{M} = \text{GEN}$, it routes the request to the scene generating module $\mathcal{G}$. 

To construct the initial scene representation, we leverage state-of-the-art text-to-video generation models. We introduce an inference-time guidance mechanism that dynamically adjusts motion trajectories while sampling from the conditional distribution: $p(z | y, \mathcal{T}, i) $, where $z$ represents the generated latent features, $y$ is the input text, and $\mathcal{T}$ specifies a predefined trajectory associated with motion-related tokens $y_n$. This adjustment ensures that generated objects move according to user-specified directions without requiring additional training. The LLM agent first parses the motion direction description from the natural language input into a structured trajectory representation and assigns corresponding bounding boxes. These bounding boxes serve as guidance inputs for the scene generator, ensuring that objects move in the specified direction and at the indicated speed within the generated 4D scene.

\noindent \textbf{Bounding Box Definition.} 
For a given translated trajectory $\mathcal{T} = \{(x_{i}, y_{i}, t_{i})\}_{i=1}^{L}$, where $L$ is the number of key points, $(x_{i}, y_{i})$ represents the spatial location at time step $t_{i}$, we define the bounding box $B_{i}$ for frame $i$ as:
$B_{i} = \{ (x, y) : |x - x_{i}| \leq \Delta_x, |y - y_{i}| \leq \Delta_y \}$,
where $\Delta_x$ and $\Delta_y$ define the spatial tolerance, determining the size of the bounding box. These values are influenced by both object size and the motion attributes extracted from the language.

\noindent \textbf{Frame-Wise Bounding Box Allocation.} 
To account for motion speed and direction, we introduce a velocity-dependent expansion factor. If the language prompt describes fast movement (e.g., "The car moves quickly to the right"), the bounding boxes are spaced farther apart between frames to reflect rapid displacement. We define the displacement vector between consecutive points as:
$v_{i} = \frac{\| (x_{i+1} - x_{i}, y_{i+1} - y_{i}) \|}{t_{i+1} - t_{i}}$,
where $v_{i}$ represents the velocity magnitude. The bounding box displacement between frames is then scaled by a velocity factor $\lambda(v_i)$:
$x_{i+1} = x_i + \lambda v_i \cdot (x_{i+1} - x_{i}), y_{i+1} = y_i + \lambda v_i \cdot (y_{i+1} - y_{i})$.
Here, $\lambda$ is a scaling hyper-parameter that increases with velocity, ensuring that high-speed moving objects specified in the text prompts receive more widely spaced bounding boxes across frames. This adaptive allocation ensures that motion dynamics described in natural language are accurately reflected in the generated scene.

% \begin{figure}[t]
%   \centering
%   \includegraphics[width=0.6\linewidth]{figures/failure.pdf}
%   \caption{\textbf{Failure cases of state-of-the-art video generation models in adhering to spatial instructions from text prompts}. The generated object motions move in the opposite direction of the specified text prompt. The first row presents videos generated by Hunyuan~\cite{kong2024hunyuanvideo}, while the second row shows results from Cosmos~\cite{nvidia2025cosmos}.}
%   \label{fig:failure}
% \end{figure}

\noindent \textbf{Dynamic Control Submodule.}  
To enforce directional control, we introduce the Dynamic Control Submodule with guidance bounding boxes, where we modify the cross-attention layers by biasing attention scores toward locations along $\mathcal{T}$. Specifically, at each sampling step $i$, we adjust the attention response for layers attending to motion-related text tokens. This is implemented using a trajectory-aligned attention weighting mechanism inspired by~\cite{he2024mojito}, which directs attention toward spatial regions prescribed by the trajectory. Figure~\ref{fig:overall} provides an overview of this modification. By dynamically altering attention weight distributions during generation, we effectively steer object placement and movement to align with input descriptions.

Controlling spatial layouts in generative models via cross-attention has been explored in 2D scenarios~\cite{prompt_to_prompt, training-free-guidance, training-layout-control}. We extend this approach by guiding attention maps to follow a predefined trajectory $\mathcal{T}$ over time. The cross-attention score $A_{u,n}$ measures the association between spatial location $u$ and text token $y_n$, with the sum over tokens constrained to one:$
\sum_{n=1}^N A_{u,n} = 1.
$To enforce alignment with $\mathcal{T}$, we bias the attention maps to concentrate within the trajectory-defined bounding boxes $B_i$ at each timestep $i$. This is achieved through a frame-specific energy function:
\begin{equation}
\begin{aligned}
\label{energy_function}
E_{i}\left(A_{i}, B_{i}, n\right) = \left(1 - \frac{\sum_{u \in B_{i}} A_{i,u,n}}{\sum_u A_{i,u,n}}\right)^2,
\end{aligned}
\end{equation}
where $A_{i,u,n}$ represents the attention score at video timestamp $i$, spatial location $u$, and text token $y_n$. Minimizing this energy function encourages the attention distribution to remain within $B_i$, effectively guiding object placement frame-by-frame. During generation, we iteratively adjust attention maps at each denoising step to minimize $E_i$ and update latent scene features like~\cite{he2024mojito}, ensuring that the model remains aligned with $\mathcal{T}$ across successive time steps.

For models adopting the DiT~\cite{dit} architecture (e.g.,~\cite{nvidia2025cosmos}), input data is represented as a latent tensor of shape $T \times C \times H \times W$, where $T$ denotes the temporal dimension. The input video data undergoes 3D patchification via a linear projection layer, which extracts non-overlapping patches of size $(p_t, p_h, p_w)$, where $p_t$ is the patch size for the temporal dimension, $p_h$ is the height, and $p_w$ is the width, and is mapped into token embeddings for the denoiser network. This transforms the latent representation into a sequence of tokens of length $\frac{T H W}{p_t p_h p_w}$, ensuring compatibility with the model's spatiotemporal processing pipeline. To integrate directional guidance into the latent feature updates, we would modify the cross-attention map based on global visual features. Since the input undergoes 3D patch embedding in subsequent processing, we first reshape the latent feature to restore its original spatial-temporal ratio before patchification. This ensures that attention modifications remain spatially coherent and correctly aligned with the scene structure before tokenization.

% This dynamic control submodule provides an effective mechanism for ensuring that generated objects follow the intended motion direction while maintaining structural consistency across frames. By integrating trajectory-aware cross-attention modifications within the model's latent space, we achieve improved control over object movement without requiring modifications to the underlying model architecture.

\noindent \textbf{Scene Reconstruction.}
Starting from an initial generated scene representation, following~\cite{zhou2025feature4x, lei2024mosca}, we build a dynamic 3D representation that can effectively support the scene editing tasks. Given the monocular representation per frame from the 2D world generator $\mathcal{I}=\left\{I_1, \ldots, I_t\right\}$, we reconstruct the underlying dynamic 3D scene with a set of
dynamic 3D Gaussians, augmented with a unified latent feature embedding that jointly distills various 2D
foundation features useful for editing. We leverage dynamic 3D reconstruction to fuse multi-view and multi-frame 2D features into a unified 3D representation. 

To achieve this, we augment existing 3D Gaussian attributes with a latent feature $\mathcal{F}$. We learn $\mathcal{F}$ along with lightweight task-specific decoders $\{\mathcal{D}^1, \ldots, \mathcal{D}^{S}\}$, where the decoder maps the latent feature $\mathcal{F} \in \mathbb{R}^D$ to the editing feature space $\mathcal{F}^{s} \in \mathbb{R}^{D_s}$. 

During optimization, we attach a feature vector $f_j \in \mathbb{R}^{D}$ to each 3D Gaussian $G_j \in \mathcal{G}$, warp $G_j$ to the target timestep $\tau$ following~\cite{lei2024mosca}, and rasterize $f_j$ using the same approach as Gaussian color $c_j$ in~\cite{zhou2024feature}. The RGB and feature reconstruction at viewpoint $v$ and timestep $\tau$ are calculated via:
\begin{equation}
\begin{aligned}
\hat{I}_{\tau}^{v} &= \operatorname{rasterize}(v, \{\operatorname{warp}(G_j, \tau), c_j)\}_{G_j \in \mathcal{G}}), 
\end{aligned}
\end{equation}
\begin{equation}
\begin{aligned}
\hat{F}_{\tau}^{v} &= \operatorname{rasterize}(v, \{\operatorname{warp}(G_j, \tau), f_j)\}_{G_j \in \mathcal{G}}).
\end{aligned}
\end{equation}
The reconstructed feature map $\hat{F}_{\tau}$ (with viewpoint $v$ omitted for brevity) is passed through the corresponding decoder $\mathcal{D}^s$ to obtain the task-specific feature representation $\hat{F^{s}_{\tau}}$, which is supervised against the ground truth feature map obtained from the 2D encoder $\mathcal{E}^{s}$. The feature loss $L_{\text{feat}}$ is optimized with the original MoSca~\cite{lei2024mosca} loss terms:
\begin{align}
L_{\text{feat}} = \sum_{s=1}^{S} \operatorname{MSE}(\hat{F^s_{\tau}}, F^s_{\tau}), 
\end{align}
\begin{align}
\hat{F^{s}_{\tau}} = \mathcal{D}^s(\hat{F_{\tau}})
\quad \quad F^s_{\tau} = \mathcal{E}^s(I_{\tau}).
\end{align}
Similar to~\cite{zhou2025feature4x},
% embedding high-dimensional features (e.g., 512-dimensional CLIP features) within explicit 3D Gaussian scene representations remains a challenge~\cite{yu2024language,shi2024language,zhou2024feature,qin2024langsplat,zuo2024fmgs,lee2024rethinking}. 
We use an MLP-based decoder trained on 2D rendered feature maps but applied directly to 3D Gaussian features during inference.

\begin{figure*}[tbp]
  \centering
%   \fbox{\rule{0pt}{3in} \rule{0.9\linewidth}{0pt}}
\includegraphics[width=\linewidth]{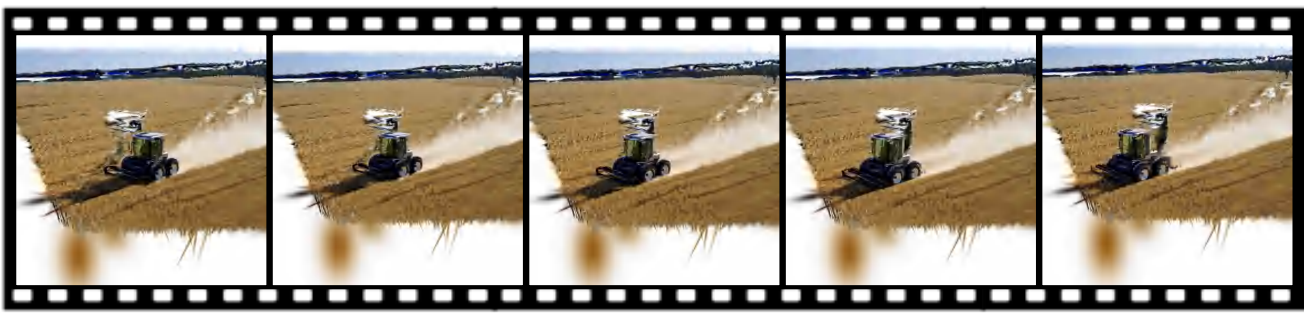}
  \caption{\textbf{Qualitative examples of 4D scene editing in \modelname for object motion control during the generation stage.} \modelname allows specifying different object motion directions in natural language forms and subsequently changes the scene to ensure objects move according to the given instructions.}
   \label{fig:motion_example}
\end{figure*}

\begin{figure*}[t]
  \centering
%   \fbox{\rule{0pt}{3in} \rule{0.9\linewidth}{0pt}}
\includegraphics[width=\linewidth]{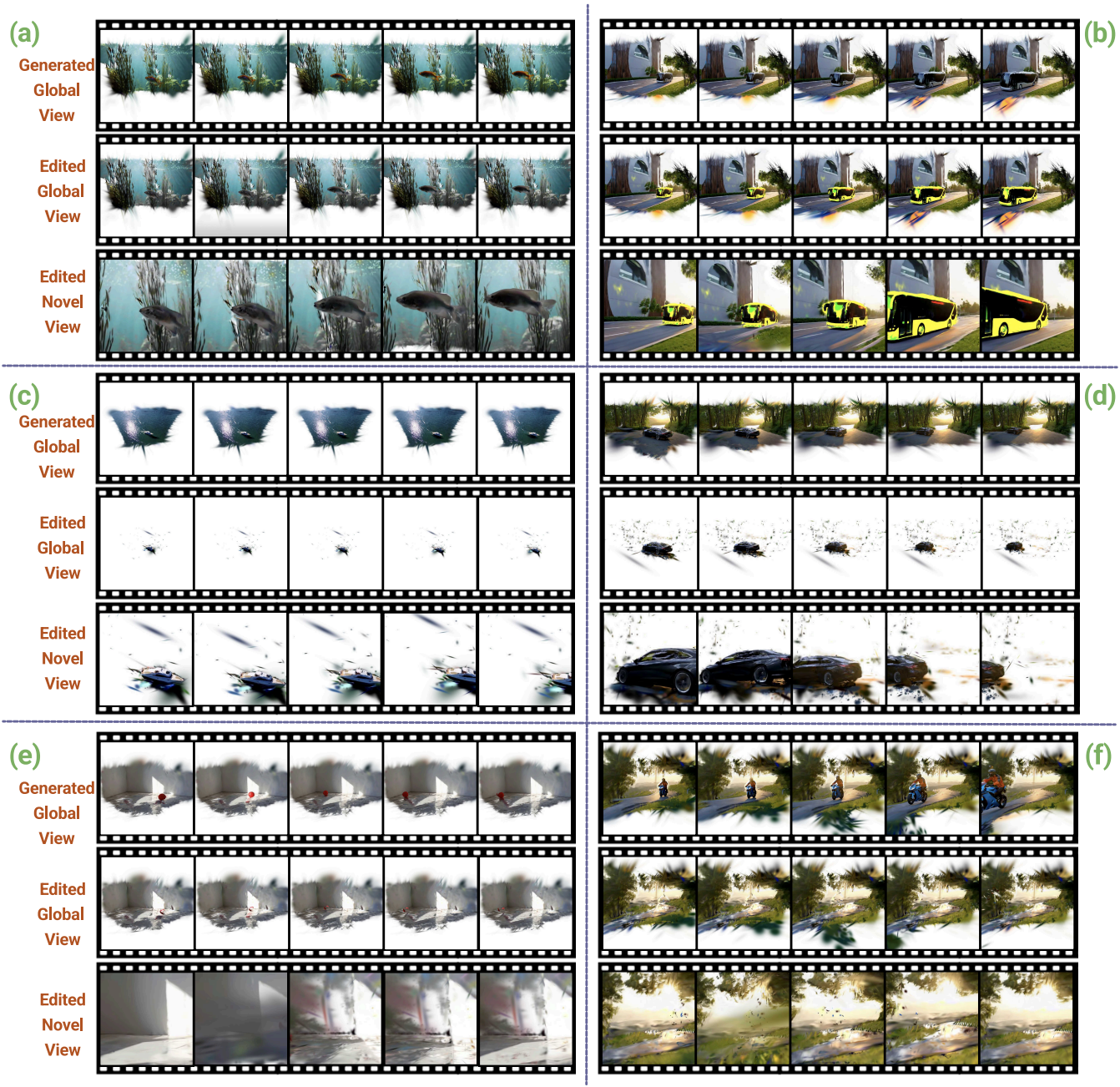}
  \caption{\textbf{Qualitative examples of 4D scene editing in \modelname during the reconstruction stage}.  
  (a) and (b) demonstrate \textit{color editing}, (c) and (d) show \textit{object extraction}, while (e) and (f) illustrate \textit{object removal}. In each subfigure: The first row shows the generated global view from the text prompt; The second row presents the global view after scene editing; The third row displays the novel view after editing.  The language commands for each example are as follows: \underline{(a)} "\textit{The fish swims through the crystal-clear waters from right to left}" to generate the scene, followed by "\textit{Make the color of the fish and seaweed black.}" \underline{(b)} "\textit{The bus is moving from right to left}" to generate the scene, followed by "\textit{Make the bus yellow.}" \underline{(c)} "\textit{A serene boat glides gracefully through tranquil waters from left to right}" to generate the scene, followed by "\textit{Extract the boat.}" \underline{(d)} "\textit{A car is moving from right to left through a serene sunlit landscape}" to generate the scene, followed by "\textit{Extract the car.}" \underline{(e)} "\textit{A small, vibrant red rubber ball is bouncing from right to left}" to generate the scene, followed by "\textit{Delete the ball.}" \underline{(f)} "\textit{A sleek black motorcycle is gliding effortlessly from right to left}" to generate the scene, followed by "\textit{Delete the motorcycle.}"}
   \label{fig:editing_example}
\end{figure*}

\begin{table*}[t]
\centering
\caption{Quantitative comparison between real-world scenes (Davis) and ~\modelname generated 4D scenes with two different base models.}
\label{tab:reconstruction_quality}
\resizebox{\textwidth}{!}{
\begin{tabular}{lccccc}
\toprule
\toprule
\textbf{Method} & \textbf{BRISQUE~\cite{brisque} $\downarrow$} & \textbf{NIQE~\cite{niqe} $\downarrow$} & \textbf{CLIP Similarity~\cite{clip} $\uparrow$} & \textbf{QAlign Quality~\cite{wu2023qalign} $\uparrow$} & \textbf{QAlign Aesthetic~\cite{wu2023qalign} $\uparrow$}\\
\midrule
\rowcolor[gray]{0.9}  \multicolumn{6}{l}{\textit{Overall Average}}\\
Davis (Real) & 31.639 & 3.551 & 0.250 & \textbf{3.432} & \textbf{2.254}\\
Backbone I & \textbf{18.380} & \textbf{3.286} & \textbf{0.263} & 3.350 & 2.114\\
Backbone II & 23.411 & 3.392 & 0.261 & 3.309 & 2.074\\
\midrule
\multicolumn{6}{l}{\textit{Example Scene Comparisons:}} \\
\midrule
\rowcolor[gray]{0.9}  \multicolumn{6}{l}{\textit{Scene: sheep}} \\
Davis (Real) & 18.090 & \textbf{2.173} & 0.266 & \textbf{4.371} & \textbf{2.891}\\
Backbone I & \textbf{9.638} & 3.454 & \textbf{0.300} & 3.623 & 1.918\\
Backbone II & 14.476 & 3.951 & 0.290 & 3.734 & 2.071\\
\midrule
\rowcolor[gray]{0.9} \multicolumn{6}{l}{\textit{Scene: snowboard}} \\
Davis (Real) & 34.904 & 3.719 & 0.269 & 2.715 & 1.991\\
Backbone I & \textbf{18.163} & \textbf{2.836} & \textbf{0.313} & \textbf{4.011} & \textbf{2.306}\\
Backbone II & 29.504 & 3.412 & 0.279 & 2.451 & 1.970\\
\midrule
\rowcolor[gray]{0.9}  \multicolumn{6}{l}{\textit{Scene: elephant}} \\
Davis (Real) & 16.815 & \textbf{2.317} & 0.284 & \textbf{4.048} & \textbf{2.764}\\
Backbone I & \textbf{15.140} & 3.171 & \textbf{0.308} & 3.634 & 2.511\\
Backbone II & 16.877 & 3.982 & 0.301 & 3.765 & 2.590\\
% \midrule
% \rowcolor[gray]{0.9}  \multicolumn{6}{l}{\textit{Scene: soapbox}} \\
% Davis (Real) & 47.308 & 4.527 & 0.221 & - & -\\
% Backbone I & \textbf{18.207} & \textbf{2.893} & \textbf{0.332} & - & -\\
% Backbone II & 38.490 & 3.795 & 0.294 & - & -\\
\bottomrule
\bottomrule
\end{tabular}}
\end{table*}

\subsection{Scene Editing Module}
If the LLM agent $\mathcal{A}$ determines $\mathcal{M} = \text{EDIT}$, it routes the request to the scene editing module $\mathcal{E}$, which modifies an existing 4D scene $\mathcal{S}$ based on the structured queries $\mathcal{Q}$. The editing operations include:
\begin{itemize}
    \item Appearance Editing: Changing the object color.
    \item Object Manipulation: Removing or extracting objects from the scene.
\end{itemize}
We further utilize the LLM agent to parse the input languages into executable commands and then perform follow-up executions, such as Color Editing, Object Removal, and Object Extraction. The modules can also be extended to support more executions.

The LLM agent is then used to optimize configuration parameters based on natural language prompts, perform precise queries, and iteratively refine results, enabling intelligent 4D scene manipulation. For instance, given a user instruction such as "Delete the ball" or "Change the ball's color to blue," the agent first parses the prompt and generates a set of configuration options with varying parameters relevant to the task. Specifically, it computes the probability $\mathbf{p}(\tau\mid j)$ of a 3D Gaussian being associated with a prompt $\tau$:
$
\mathbf{p}(\tau \mid j) = \frac{\exp (s)}{\sum_{s_i \in \mathcal{T}} \exp (s_i)},
$
where $s$ is the cosine similarity between the semantic feature $f_j$ of the 3D Gaussian and the query feature $
q(\tau): s = \frac{f_j \cdot q(\tau)}{\|f_j\|\|q(\tau)\|}.
$
The LLM agent then iterates over different threshold values to filter Gaussians with low probability scores and generate sample images using the 4D feature field, evaluating which configuration best aligns with the intended edit.

Once the optimal configuration is determined, the LLM agent $\mathcal{A}$ applies the selected parameters consistently across all frames in the video sequence, ensuring coherence in dynamic 4D scene editing. For instance, when modifying an object's color, the LLM agent iteratively adjusts the threshold to isolate and modify only the target object while preserving the rest of the scene. This process continues until either the input maximum iteration count parameter is reached or the threshold falls below a specified limit, ensuring precise and controlled edits.

\section{Experiments}  
\subsection{Datasets}  
We evaluate \modelname by comparing its generated 4D scenes against real-world videos from the DAVIS dataset~\cite{davis2017}. Specifically, we construct textual prompts from DAVIS annotations to generate corresponding 4D scenes and compare with reconstructed 4D scenes from DAVIS videos, assessing their quality relative to real-world counterparts.

\subsection{Quantitative Results on Generated Scene Quality}
We evaluate the quality of the reconstructed 4D scenes using four metrics:  BRISQUE (Blind/Referenceless Image Spatial Quality Evaluator)~\cite{brisque} is a no-reference image quality assessment metric that measures perceptual distortions based on natural scene statistics; NIQE (Natural Image Quality Evaluator)~\cite{niqe} is another no-reference quality metric that evaluates the deviation of an image from learned natural scene statistics; CLIP Similarity~\cite{clip} quantifies the semantic consistency between generated and real-world scenes by computing the cosine similarity between image embeddings extracted from the CLIP model; QAlign~\cite{wu2023qalign} is the current state-of-the-art method for image quality assessment, leveraging a large multimodal model fine-tuned on publicly available image quality assessment datasets. 

The results are summarized in Table~\ref{tab:reconstruction_quality}, where we compare our method on two video generation backbones with real-world scenes (Davis).  Backbone I is modified from CogVideoX~\cite{hong2022cogvideo}, and Backbone II is modified from Cosmos~\cite{nvidia2025cosmos} to achieve spatial control and 4D scene generation. The results demonstrate that our generated 4D scenes achieve comparable or better quality than real-world scenes using both two backbones, with Backbone I showing significantly better BRISQUE scores, slightly better NIQE scores, and improved CLIP similarity. This improvement can be attributed to the quality of~\modelname generated 4D scenes, as well as the fact that~\modelname generated camera views remain fixed throughout, resulting in higher metric scores, whereas real-world video sequences typically involve dynamic camera movement, which can introduce additional variance in evaluation.

\subsection{Qualitative Results}
We present qualitative results demonstrating the capabilities of our method in 4D scene generation and editing. As shown in Figure~\ref{fig:editing_example}, our approach generates 4D scenes with realism comparable to real-world environments. Additionally, it enables dynamic object motion editing, allowing objects to be manipulated to move in different directions based on user instructions (Figure~\ref{fig:motion_example}). Beyond motion control, our method supports appearance modifications, such as color editing (Figure~\ref{fig:editing_example}), ensuring consistent alterations across frames. Moreover, it facilitates structural modifications, including object extraction, which isolates objects without disrupting the surrounding scene, and object removal, which eliminates target objects while preserving background coherence. These results highlight the versatility of our approach in controlling, modifying, and refining 4D scene generation.

\section{Conclusion}
We introduced \modelname, a framework for language-guided generation and interactive editing of 4D scenes. By combining large language models with trajectory-guided generation and feature field distillation, \modelname produces dynamic, view-consistent environments that can be modified at the object level through natural language commands. We believe that such simulation tools can accelerate progress in robot learning and provide a flexible platform for perception, planning, and interaction.

\bibliographystyle{unsrt}
\bibliography{main}

\end{document}